\documentclass[conference]{IEEEtran}
\IEEEoverridecommandlockouts

\usepackage{cite}
\usepackage{amsmath,amssymb,amsfonts}
\usepackage{algorithmic}
\usepackage{graphicx}
\usepackage{textcomp}
\usepackage{xcolor}
\usepackage{url}
\def\BibTeX{{\rm B\kern-.05em{\sc i\kern-.025em b}\kern-.08em
    T\kern-.1667em\lower.7ex\hbox{E}\kern-.125emX}}
\begin{document}

\definecolor{gray}{rgb}{0.66, 0.66, 0.66}

\title{Interpretable time series neural representation for classification purposes}

\author{\IEEEauthorblockN{Etienne Le Naour}
\IEEEauthorblockA{\textit{Sorbonne Université, EDF R$\&$D}\\
Paris, France \\
etienne.le-naour@edf.fr}
\and
\IEEEauthorblockN{Ghislain Agoua}
\IEEEauthorblockA{\textit{EDF R$\&$D}\\
Palaiseau, France\\
ghislain.agoua@edf.fr}
\and
\IEEEauthorblockN{Nicolas Baskiotis}
\IEEEauthorblockA{\textit{Sorbonne Université}\\
Paris, France\\
nicolas.baskiotis@isir.upmc.fr}
\and

\IEEEauthorblockN{Vincent Guigue}
\IEEEauthorblockA{\textit{AgroParisTech}\\
Palaiseau, France\\
vincent.guigue@isir.upmc.fr}
}

\maketitle

\begin{abstract}
Deep learning has made significant advances in creating efficient representations of time series data by automatically identifying complex patterns. However, these approaches lack interpretability, as the time series is transformed into a latent vector that is not easily interpretable. On the other hand, Symbolic Aggregate approximation (SAX) methods allow the creation of symbolic representations that can be interpreted but do not capture complex patterns effectively. In this work, we propose a set of requirements for a neural representation of univariate time series to be interpretable. We propose a new unsupervised neural architecture that meets these requirements. The proposed model produces consistent, discrete, interpretable, and visualizable representations. The model is learned independently of any downstream tasks in an unsupervised setting to ensure robustness. As a demonstration of the effectiveness of the proposed model, we propose experiments on classification tasks using UCR archive datasets. The obtained results are extensively compared to other interpretable models and state-of-the-art neural representation learning models. The experiments show that the proposed model yields, on average better results than other interpretable approaches on multiple datasets. We also present qualitative experiments to asses the interpretability of the approach.
\end{abstract}


\section{Introduction}

Unsupervised representation learning approaches aim to build representation for time series by capturing their underlying distribution without expert knowledge or human supervision. They have demonstrated good performances for clustering \cite{ma2019learning}, classification \cite{franceschi2019unsupervised, malhotra2017timenet, DBLP:conf/aaai/YueWDYHTX22}, missing values imputation or forecasting \cite{zerveas2021transformer}. Despite these good performances for downstream tasks, the neural representations models in the literature lack interpretability. In \cite{DBLP:journals/pami/BengioCV13} a review of representation learning, the authors emphasize that good representations should have the ability to extract \textit{Explanatory Factors} and should guarantee \textit{Temporal Consistency}. Current approaches do not meet these criteria. Indeed, for most existing approaches \cite{franceschi2019unsupervised, DBLP:conf/aaai/YueWDYHTX22, DBLP:journals/corr/abs-2206-08496}, the representation results from mapping signals to a latent vector with no temporal consistency and in which weights have no meaning. These representations fail to provide interpretability when used for downstream tasks like classification, which is problematic for critical decision-making.

\begin{figure}[!htb]
    \centering
    \includegraphics[width=.85\linewidth]{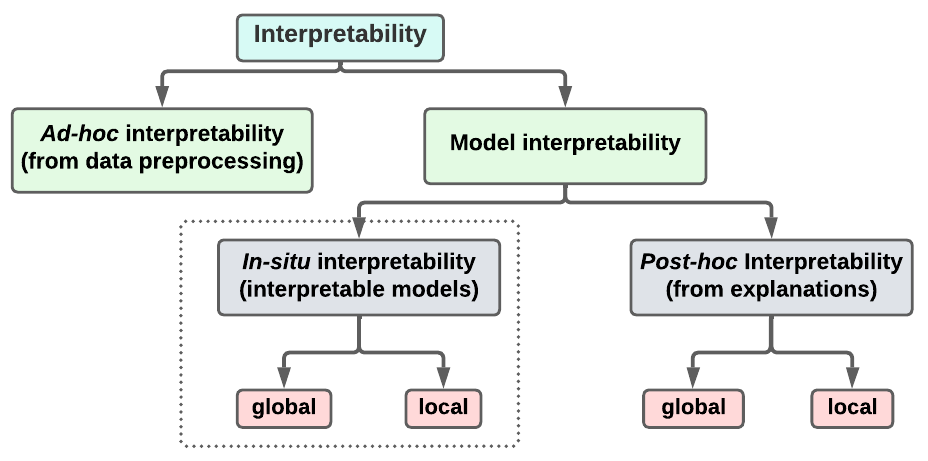}
    \caption{Interpretable time series model taxonomy introduced in \cite{DBLP:phd/hal/Wang21c}}
    \label{fig:taxonomy}
\end{figure}

However, interpretability is a concept that is not universally agreed upon \cite{DBLP:journals/cacm/Lipton18}, with confusion arising from the different meanings of interpretability and explicability. For time series models,  \cite{DBLP:phd/hal/Wang21c} offers a clear taxonomy of the interpretability shown in Figure \ref{fig:taxonomy}. Post-hoc interpretability refers to methods that analyze the model after the training and are generally model-agnostic. It is often related to the eXplainable Artificial Intelligence (XAI) research field. However, post-hoc methods only explain the decision for a specific instance (or specific features), and additional methods are required to understand the overall model.  

On the other hand, in-situ interpretable models are self interpretable. The interpretability arises directly from the model without any other process being applied after the training phase \cite{DBLP:journals/natmi/Rudin19, DBLP:phd/hal/Wang21c}. The level of this interpretability can be local or global. In \cite{DBLP:journals/cacm/Lipton18}, global interpretability is defined as a model that is easy for a human to understand and requires low computational complexity. In contrast, local interpretability is a way to interpret a model's decision for a particular instance. Global interpretability often implies local interpretability, but the reverse is not true.

In this work, we focus on global in-situ interpretability rather than post-hoc explainability, as global in-situ models are inherently interpretable and can be understood both for individual instances and the model as a whole. This paper aims to develop a global in-situ interpretable neural method for time series representation. The first contribution of this work is to define the requirement to bridge the gap between symbolic representation and neural representation to ensure a global interpretable neural symbolic representation for time series data. Indeed, most successful interpretable models come from symbolic machine learning, such as symbolic aggregate approximation (SAX), which creates interpretable symbolic representations of time series data. However, the information captured by these symbols is limited and does not provide global interpretability of the representation. On the other hand, neural representation learning methods achieve great performances but are definitively not in-situ interpretable. Section \ref{Requirements} analyses the criteria that must be respected to guarantee global in-situ interpretability.

We propose a novel unsupervised neural network that fills these requirements in Section \ref{sec:model}. The neural network is based on an auto-encoder architecture and vector quantization mechanism \cite{gray1984vector, van2017neural, fortuin2018som}. The unsupervised setting is a crucial choice for the generalization of the learned representation. Moreover, it allows to re-use the extracted representation for several classification tasks.


To demonstrate the qualities of the proposed architecture, Section \ref{Classif_section} presents an application of our learned symbolic neural representation to classification tasks. A  simple linear classifier over the interpretable symbolic representations is used to solve classification tasks efficiently. The linearity of the classifier preserves interpretability in the representation, providing both global and local interpretability for understanding the decision made by the classifier. 



Our main contributions can be summarized as follows:

\begin{itemize}
    \item We define and formalize the fundamental requirements to construct interpretable symbolic neural representations for time series.
    \item We propose an unsupervised neural network architecture that satisfies the above requirements. 
    \item We use these representations for downstream classification tasks (while preserving interpretability of the representation) and evaluate them through quantitative experiments on the UCR archive. 
    \item We provide qualitative experiments to capture local and global interpretability.
\end{itemize}
\section{Related content}
\label{RelatedContent}

Constructing time series representations is a fundamental challenge that can be performed unsupervised or based on a specific task. This related content presents both neural representation methods, which are not easy to interpret, and classical representation methods, which are interpretable. In most cases, classical representation methods are combined with a classification task.

\paragraph{Unsupervised neural representation learning for time series classification} In recent works, models of neural representations of time series have emerged. These models typically learn the representation in an unsupervised way and then use the learned representation for a specific task in a second step. In \cite{malhotra2017timenet}, the authors use a deep unsupervised representation to solve a subsequent classification task for time series. They used an architecture composed of Recurrent Neural Networks to build a representation that would later be used for classification. In \cite{franceschi2019unsupervised}, authors build a time series representation using a convolutional encoder and a contrastive loss \cite{mikolov2013efficient}. The representation space brings together series (and sub-series) that are similar. Afterward, a support vector machine (SVM) is applied on top of the representation to solve the classification problem. Then, several papers attempted to construct vector representations of time series using contrastive loss \cite{DBLP:conf/aaai/YueWDYHTX22, DBLP:journals/corr/abs-2206-08496}. Some recent works \cite{zerveas2021transformer} have tried to build an unsupervised representation of time series using transformers mechanisms \cite{vaswani2017attention} inside an auto-encoder. These neural representations can capture a lot of information, and the downstream tasks learned from them are very efficient. However, the representations and thus the downstream tasks cannot be interpreted with these models.

\paragraph{Attempt to construct an interpretable neural representation for time series}  Recently, progress has been made in making neural representations of time series interpretable. In \cite{DBLP:conf/kdd/LiCZDNZCH22}, the authors have attempted to decompose time series into disentangled semantic factors (for both individual factors and group segment factors) using Variational Auto-Encoder (VAE), LSTM, and a disentanglement strategy. The limitation of disentangled representations is that it is difficult to assess the disentanglement of latent factors when the initial semantic properties of the time series are unknown. This unsupervised representation is interesting for generation, but difficult to adapt for classification.  Other work, such as \cite{DBLP:conf/kdd/Luo0LLZ022}, has attempted to obtain interpretable differential operators from multivariate time series. However, they are specific to forecasting.

Although neural methods for representing time series are relatively new, non neural interpretable methods for representing time series are widely studied.

\paragraph{The Symbolic Aggregate approximation (SAX)} SAX methods \cite{lin2003symbolic, lin2007experiencing} create a symbolic representation of a time series by combining local statistics, which are calculated by taking the average of different segments of the time series. Each average is assigned a symbol based on its value. A sequential symbolic representation of the original time series is formed by mapping these local averages to symbols. Multiple SAX representations can be obtained by taking more or less local averages. SAX symbolic elements operate in the time domain and are interpretable, unlike the Symbolic Fourier approximation (SFA) method \cite{schafer2012sfa}, which operates in the frequency domain. In \cite{lin2003symbolic}, the authors introduce the SAX method as a versatile unsupervised approach for various downstream tasks, but in practice, SAX methods are typically used in combination with a classifier on top of the representation.
There are several methods to classify on top of SAX representations. For example, the \textit{SAX-VSM} method \cite{Senin2013} is based on the frequency of subsequences within each class. The \textit{SAX-SEQL} method \cite{le2017time} searches the entire space of subsequences for the most discriminating subsequences using logistic regression based on coordinate descent.
SAX methods have proven useful for making interpretable classification decisions from the constructed symbolic representation. However, this interpretability is only local. Indeed, when we obtain a discriminating symbolic sub-sequence, we can highlight the corresponding sub-part of the time series. Nevertheless, we cannot reconstruct what the model has learned because several sub-series can give the same symbolic sub-sequence. Additionally, as stated in \cite{DBLP:journals/cacm/Lipton18}, when the number of representations used for classification is excessive, such as in the case of multi-SAX-SEQL \cite{le2019interpretable}, it becomes difficult to interpret the classification results.

\paragraph{Shapelet methods} These methods aim to find the sub-sequences of the time series that most discriminate between classes \cite{DBLP:journals/datamine/YeK11}. This representation method is supervised because it relies on a downstream task. In addition, the resulting representations are partial, since only the shapelets that discriminate between classes are extracted at the end of the process. Once the optimal shapelets are found, if the classifier applied on top of them is interpretable (linear regression or decision tree), the whole process is interpretable at both local and global levels. However, the original method is costly because it is necessary to search the whole space for possible subsequences. In \textit{Fast Shapelets} (FS) \cite{DBLP:conf/sdm/KeoghR13}, the authors proposed to speed up the discovery of discriminative subsequences by using a SAX representation to discover the subparts of the series where shapelets should be searched. Although the method speeds up the discovery of discriminative shapelets, it remains computationally expensive and the accuracy could be better. To avoid these problems, in \cite{DBLP:conf/kdd/GrabockaSWS14}, authors propose the \textit{Learning Shapelets} (LTS) method. The goal is to learn the discriminating shapelets rather than to search for them in the whole space of possibilities. This method has led to improvements in accuracy. However, it generates a large number of shapelets that are almost the same. This negatively affects interpretability. Afterward, methods have been proposed to learn a limited number of shapelets and thus reinforce the interpretability of the model. In \cite{DBLP:conf/ictai/WangEFMT20}, authors propose learning a small number of discriminating shapelets by training a Generative Advertial Network (GAN) and a classifier.

To the best of our knowledge, there is no unsupervised learning method capable of learning an interpretable neural representation for time series. In the next section, we set out the requirements for constructing interpretable symbolic neural representations. These criteria will link traditional methods of representing time series using symbols and unsupervised neural representation methods.
\section{Requirements for an interpretable symbolic neural representation}
\label{Requirements}

In order to ensure that a neural representation of a time series is interpretable, we set out several requirements that we consider essential.

For this purpose, we introduce some notation for this section. We consider that we have a dataset of $N$ samples. For an instance $i$ ($i \in \{1,...,N\}$), the univariate time series is denoted by the vector $\boldsymbol{x_{i}}$ of length $T$: $\boldsymbol{x_i} =\left(x_{i,1}, \ldots, x_{i,T}\right) \in \mathbb{R}^{T}$. Let $\boldsymbol{r_i}$ be the symbolic neural representation composed of $T^{'}$ elements for $\boldsymbol{x_i}$. We denote by $\mathbb{A}$ the support (alphabet) common to all these elements: $\boldsymbol{r_i} =(r_{i,1}, \ldots, r_{i,T^{'}}) \in \mathbb{A}^{T'}$. Then $\phi_{\theta}$ is the function that maps the time series into the representation, and $\psi_{\theta^{'}}$ is the function that goes from the representation to the reconstruction space of the time series. To simplify the reading, we omit the indices $i$ for the vectors $\boldsymbol{r}$ and $\boldsymbol{x}$.

\paragraph{Requirement n°1 - discrete symbolic representation} The purpose of a symbolic neural representation method is to capture complex phenomena within the representation (as neural representations do) while being able to interpret and visualize the representation elements (as symbolic representations tend to do). Thus, the support $\mathbb{A}$ of each element must be discrete and limited (e.g. $Card(\mathbb{A}) = 32$). In addition, the support must be common to all elements of the symbolic representation. This limits the number of possible patterns. Once we obtain the symbolic representation $\boldsymbol{r}$, we can use the classifiers used in the dictionary methods \cite{schafer2015boss, le2017time, le2019interpretable} (see Section \ref{Classif_section}).

\begin{figure}[!htb]
    \centering
    \includegraphics[width=.70\linewidth]{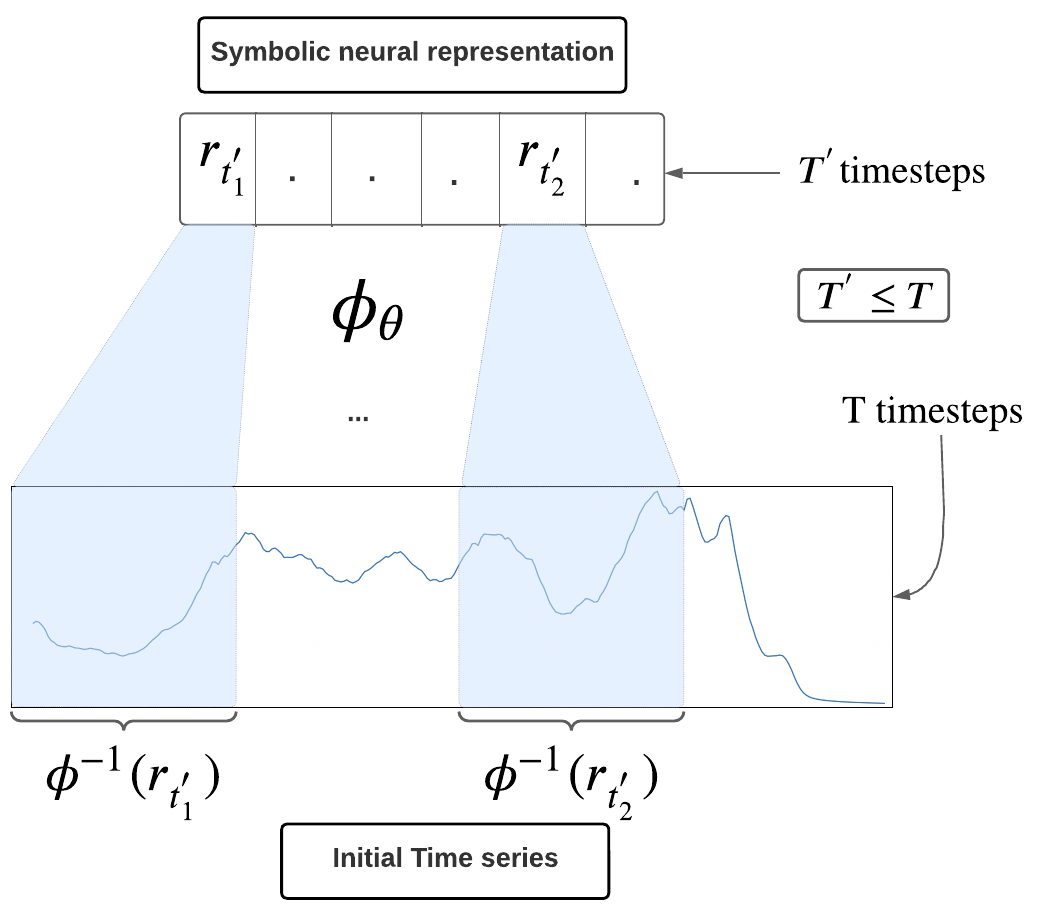}
    \caption{Temporal consistency visualization. \textit{Coffee dataset}}
    \label{fig:tempo_consistency}
\end{figure}

\paragraph{Requirement n°2 - temporal consistency} For a time series $\boldsymbol{x}$ of length $T$, learning a contracted representation $\boldsymbol{r}$ of length $T^{'}$ will mechanically lead to a contraction of the time dimension ($T^{'} < T$). We then define temporal consistency by two properties. First, each element of the representation is a function of a portion of the original time series. Thus, for each element of the representation, we must be able to compute the pre-image of the element. Second, the representation must preserve the original temporal order despite the contraction of the temporal dimension. To illustrate this property, consider the case where $r_{t^{'}_{1}}$ is an element of the representation and $r_{t^{'}_{2}}$ is another element of the representation that occurs after $r_{t^{'}_{1}}$. As shown in Figure \ref{fig:tempo_consistency}, the temporal consistency of the representation ensures that the pre-image of $r_{t^{'}_{1}}$  must precede the pre-image of $r_{t^{'}_{2}}$ in time. 

\begin{figure}[htbp]
    \centering
    \includegraphics[width=.80\linewidth]{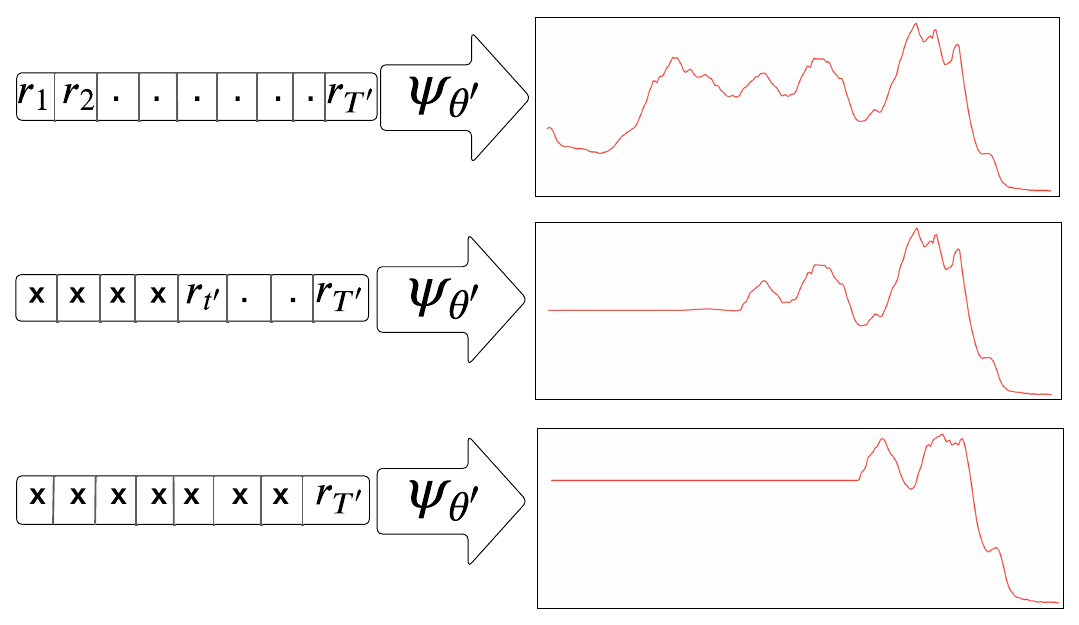}
    \caption{Visualization of how the representation is decoded for the complete representation, as well as for half of the representation and for only the last element. \textit{Coffee dataset}.}
    \label{fig:decodable}
\end{figure}

\paragraph{Requirement n°3 - a decodable representation} Being able to visualize the portion of the time series related to a specific element of the representation is important, but it is also important to be able to see what the model has learned overall. As shown in Figure \ref{fig:decodable}, $\psi_{\theta^{'}}$ should be able to reconstruct the entire time series, as well as specific parts of it, while maintaining temporal consistency.

\begin{figure}[!htb]
    \centering
    \includegraphics[width=.99\linewidth]{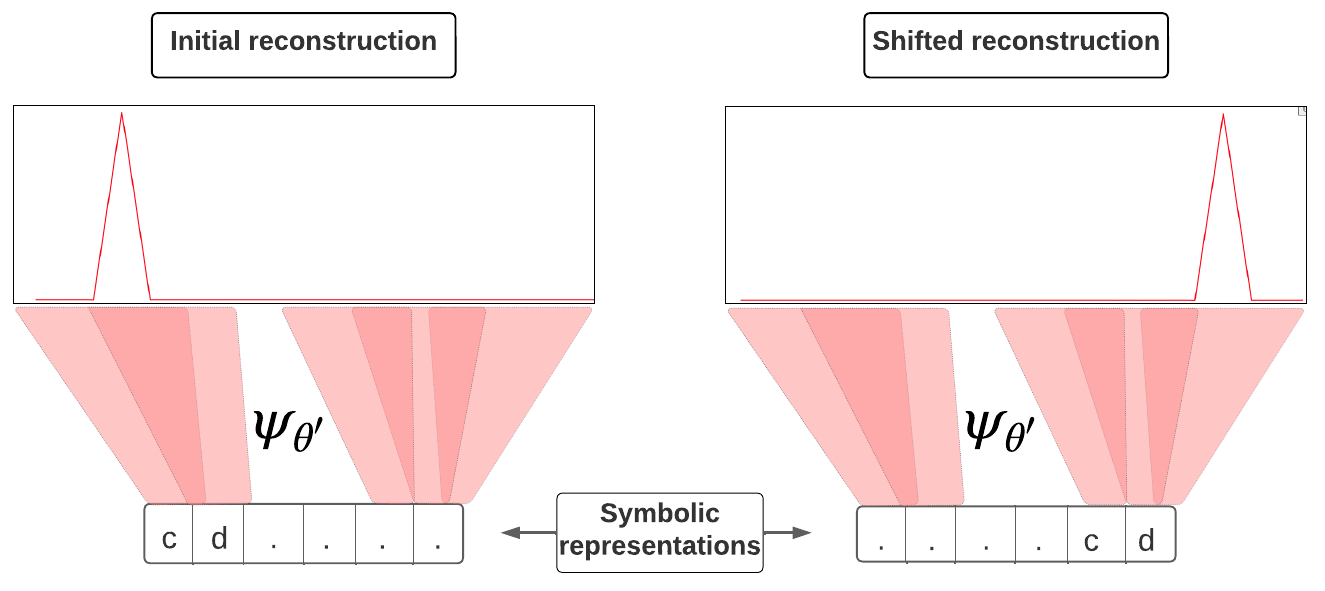}
    \caption{Shift equivariance of $\psi_{\theta'}$. The subsequence \textit{c d} appear at two different locations in the representation: $\psi_{\theta'}$ should decode the same pattern with a shift in time.}
    \label{fig:decoder_shift}
\end{figure}

\paragraph{Requirement n°4 - shift equivariance properties} We want the shift equivariance property for both $\phi_{\theta}$ and $\psi_{\theta'}$. The $\phi_{\theta}$ shift equivariance means that two patterns in the initial time series that are identical but do not occur at the same time should be encoded with the same value but not at the same place in the representation. The $\psi_{\theta'}$ shift equivariance means that two elements of the representation that have the same value but do not occur at the same time should represent the same pattern when decoded (with a time shift). This property is essential to interpret the representation elements and ensure that the same value in the representation, regardless of its position, represents the same pattern. We illustrated the shift equivariance property for $\psi_{\theta'}$ in Figure \ref{fig:decoder_shift}.

Let's formally define the shift equivariance property. Let $S$ be an element of a sequence (a single element or a subsequence), and $G_{T}$ the group of discrete translations along the temporal axis. If we take $\tau$ to be any discrete translation in $G_{T}$ and $f$ to be a function equivariant by discrete translation for $G_{T}$, then there exists $\tau' \in G_{T'}$ such that: 
$f(\tau(S)) = \tau'(f(S))$. 

\begin{figure}[!htb]
    \centering
    \includegraphics[width=.95\linewidth]{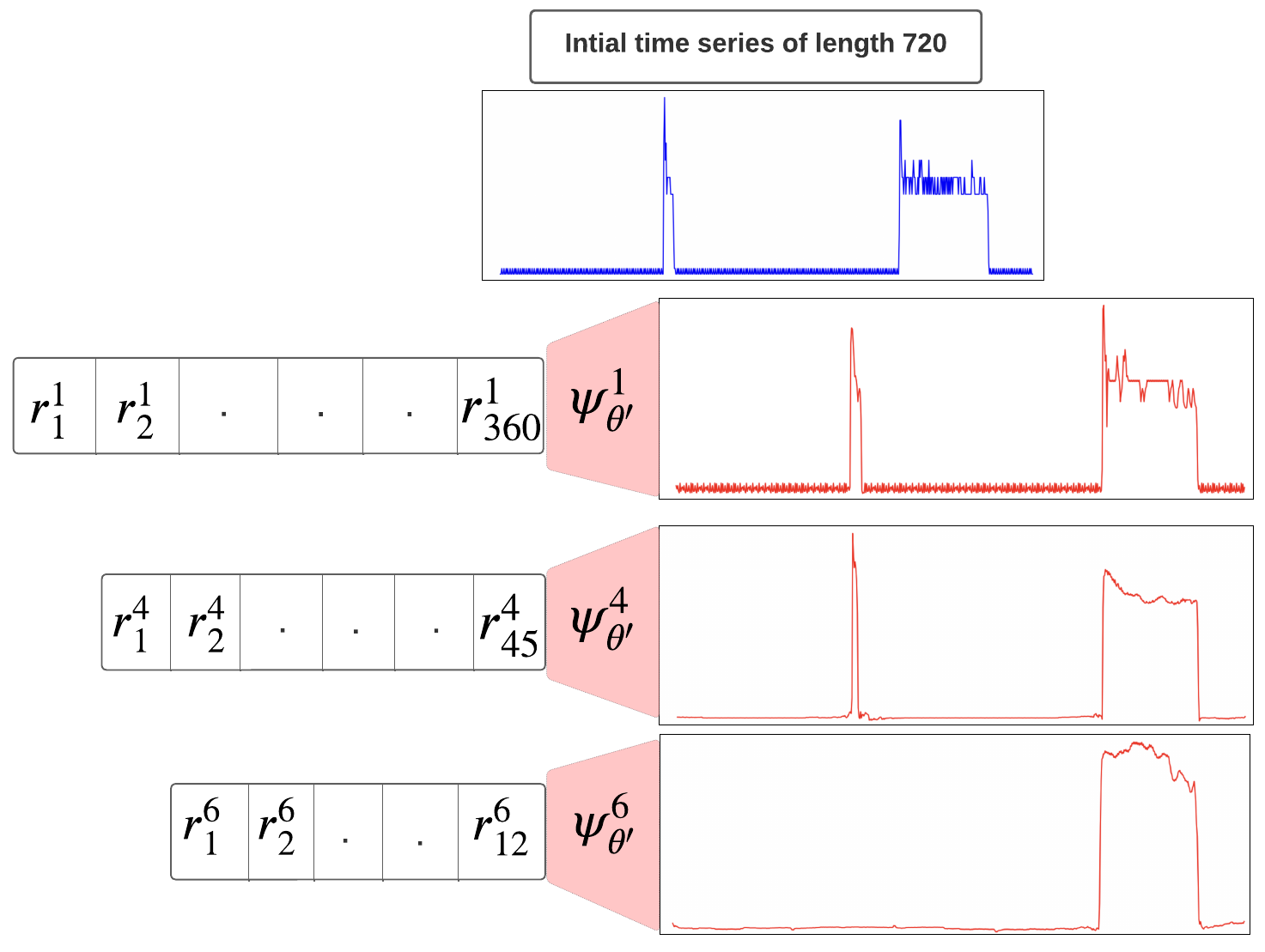}
    \caption{Visualization of different reconstructions for different level of representation. We can see that the deeper the architecture, the shorter the representation and the smoother the reconstruction. \textit{Computers dataset.}}
    \label{fig:depth}
\end{figure}

\paragraph{Requirement n°5 - a representation adjustable to the frequency level} Like an image that cannot be interpreted at the pixel level, a time series is difficult to interpret at the point level. This requirement aims to control the amount of information captured when creating the representation. For the representation to be easily understood, it is crucial to capture the appropriate frequency levels that define the time series. As illustrated in Figure \ref{fig:depth}, the depth of the representation determines whether it focuses on lower or higher frequency features, which in turn affects the length of the representation.

\section{Model}
\label{sec:model}
This section proposes an unsupervised model that respects the different requirements for building an interpretable symbolic neural representation.

\subsection{Architecture}
\label{archi}

The proposed unsupervised model architecture consists of an encoder decoder structure with a discretization mechanism within the representation space. First, the time series $\boldsymbol{x}$ is given as input to the encoder. Second, the output of the encoder is discretized using the vector quantization mechanism. This step allows us to obtain $\boldsymbol{r}$ after the learning process. Finally, the discretized elements are passed to the decoder, which returns a reconstruction of the time series $\boldsymbol{\hat{x}}$. 

\begin{figure*}[!htb]
    \centering
    \includegraphics[width=.76\linewidth]{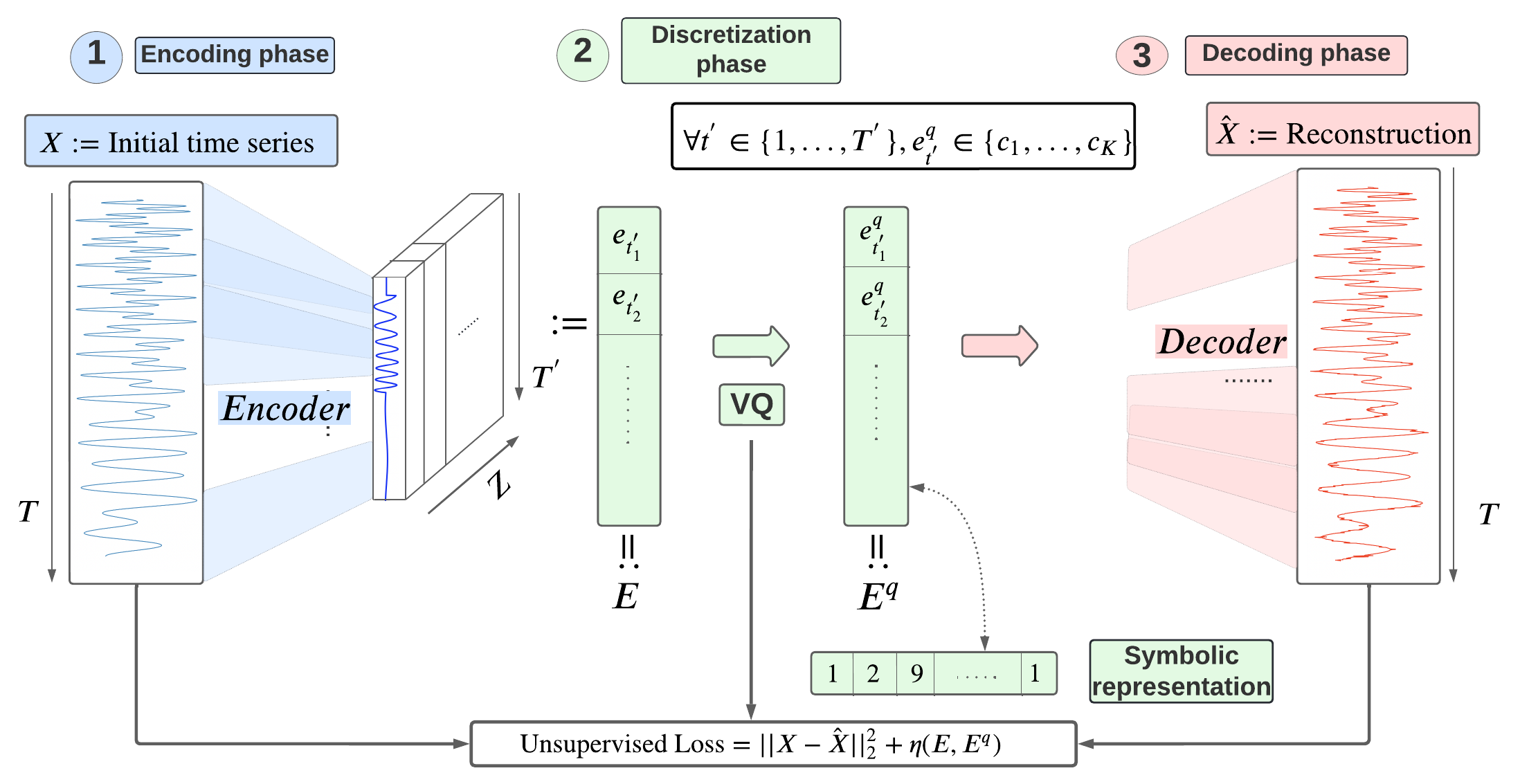}
    \caption{Global unsupervised architecture overview. $\eta$ refers the last two terms in the loss eq. \ref{total_loss}.}
    \label{fig:global-archi}
\end{figure*}

The architecture takes elements from the Vector Quantization Variationnal Auto-Encoder (VQ-VAE) \cite{van2017neural}, which was the first model to establish a latent space of discrete representation within an auto-encoder. However, we add constraints to the architecture and remove the variational part in order to meet the requirements defined in Section \ref{Requirements}. The different parts of the architecture are described below and Figure \ref{fig:global-archi} illustrates the global unsupervised architecture. Let us look at the architecture in detail, starting with the encoder.

\begin{figure}[!htb]
    \centering
    \includegraphics[width=8.5cm]{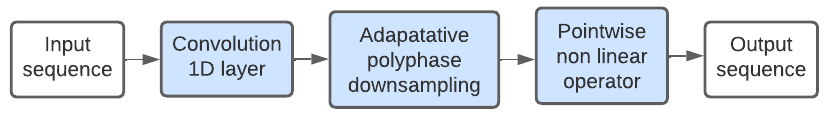}
    \caption{Inside an encoder block.}{}
    \label{EncoderBlock}
\end{figure}

\paragraph{Encoder} The encoder can be divided into a sequence of consecutive blocks with the same structure. As shown in Figure \ref{EncoderBlock}, a block consists in three operations: a 1D convolution layer, a downsampling operation, and a non-linear activation function.

The convolution layer is non-strided and has a dilation factor of zero. The input sequence is padded with a zero on each side, and the kernel size is three. Therefore, the convolution layer's output sequence length remains unchanged. Subsequently, we perform a downsampling operation on the sequence. This operation aims to reduce the sequence length by two while keeping the shift equivariance property. For this purpose, the adaptive polyphase downsampling (APS-D) proposed in \cite{DBLP:conf/acssc/ChamanD21} is used. 

To explain the APS-D operation, let us introduce $s(t)$ the convolutions output sequence of length $T_{0}$. We can then define the two sub-sequences $s_{0}$ and $s_{1}$ such that $s_{0}(t) = s(2t)$ and
$s_{1}(t) = s(2t+1)$. The APS-D operation $D^{A}_{2}$ consists in sub-sampling the sub-sequence $s_l$ such that: $$D^{A}_2(s) = s_l \quad \text{where} \quad l = \text{{arg~max}}{||s_j||_{1}} ~ (\text{for} ~ j \in \{0,1\}) $$

Now, when the downsampling process is applied to the sequence, the length of the sequence is reduced by half. Finally, a non-linear pointwise operation is applied. In practice, this operation is the LeakyReLU.

All blocks inside the encoder have the same structure. Consider an encoder with $B$ blocks and convolutions with $Z$ channels. Then the initial time series is projected into the representation space on a sequence of $T^{'} (T' \approx T / {2^{B}})$ time steps, where each point of the sequence is a vector of size $Z$.

\paragraph{Decoder} Now we can define the decoder blocks symmetrically to the encoder. The decoder is a function that projects a sequence of $T'$  vectors of size $Z$ into a reconstructed time series $\boldsymbol{\hat{x_i}}$ of size $T$. If the encoder has $B$ blocks, then the decoder will have $B$ blocks. However, the structure of a decoder block is slightly different. First, an upsampling operation (adaptive polyphase upsampling) increases the length of the sequence by two. In practice, the values of the input sequence are re-used and 0's are inserted between each value to double the size of the sequence. The 0's are inserted in an even or odd manner according to the subsequence extracted in the corresponding encoder block during the downsampling phase (if $l$ was $0$ or $1$). Next, a 1D non-strided convolutional layer with a kernel size of 3 is applied. Finally, a nonlinear pointwise operation is applied (except in the last block).

\paragraph{Discretization mechanism} Now that the encoder and decoder are defined, we need to specify the discretization method used within the representation space. After passing the time series to the encoder and thus obtaining a sequence of vectors (sequence of embeddings: $(\boldsymbol{e_{1}}, \boldsymbol{e_{2}}, ..., \boldsymbol{e_{T'}})$), we use the vector quantization (VQ) mechanism \cite{gray1984vector, van2017neural}. Given a set of $K$ centroids $\{\boldsymbol{c_j} \in \mathbb{R}^{Z},~j \in {1,..., K}\}$, VQ consists of assigning an encoder output point $\boldsymbol{e_{t'}} \in \mathbb{R}^{Z}$ to the nearest centroid:

\begin{equation}
\boldsymbol{e^{q}_{t'}} \leftarrow \boldsymbol{c_{k}} \quad \text{where} \quad k= \underset{j}{\text{{arg~min}}} ||\boldsymbol{e_{t'}}- \boldsymbol{c_{j}}||_{2}^{2}.
\label{vq_loss}
\end{equation}

\begin{figure}[!htb]
    \centering
    \includegraphics[width=0.90\linewidth]{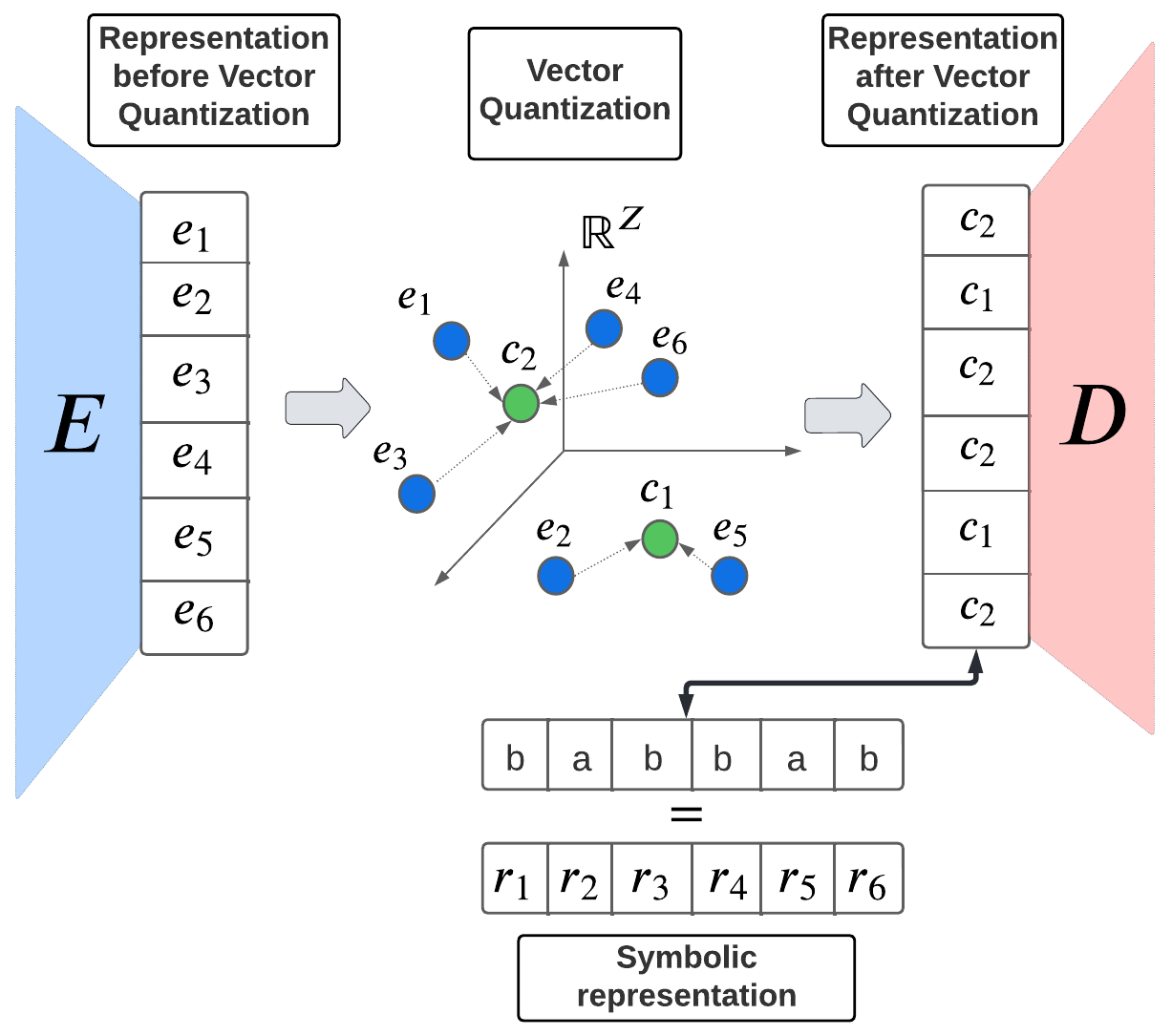}
    \caption{Visualization of the vector quantization mechanism in the context of an auto-encoder. E and D stand respectively for Encoder and Decoder.}
    \label{fig:VQ}
\end{figure}

In this way, the sequence of embedding vectors is transformed into a sequence of quantized vectors (each quantized vector has only K possible values). This mechanism is illustrated in the Figure \ref{fig:VQ}. Then, the sequence of quantized vectors ($\boldsymbol{e^{q}_{1}}$, $\boldsymbol{e^{q}_{2}}$, ..., $\boldsymbol{e^{q}_{T'}}$) is given as input to the decoder. We describe how centroids are moving through epochs in the training subsection.

\subsection{An architecture that meets the requirements}

The VQ mechanism satisfies the requirement for a discrete symbolic representation. The VQ mechanism, also known as a self-organizing map \cite{kohonen1990self}, allows the centroids to move as iterations progress. The vector quantization mechanism assigns nearby vectors to the same centroid while maintaining temporal consistency. Thus, our architecture encourages similar patterns to be quantized to the same vector. The symbolic representation is obtained directly from the centroid indices used during vector quantization. Since the index represents the vector it characterizes, the sequence of centroid indices chosen during the vector quantization phase serves as an excellent symbolic representation of the time series (thus $Card(\mathbb{A}) = K$).

The proposed architecture uses convolution operations that guarantee the temporal consistency of the representation. We can easily compute the input that produced a given output from a convolution sequence by determining the receptive field \cite{araujo2019computing}. In addition, convolutions are local operations that slide over the input, which allows them to preserve the temporal order of the input. Using a pointwise VQ mechanism does not affect the temporal consistency property of our architecture.

The symmetric and unbiased encoder/decoder design guarantees that the representation is decodable. Our loss function (eq. \ref{total_loss}) ensures that the reconstruction converges towards the original time series. It enables the visualization of the learned representation while maintaining temporal consistency and enabling separate decoding of each representation component.

Due to the adaptive polyphase upsampling/downsampling \cite{DBLP:conf/acssc/ChamanD21}, both encoder and decoder are shift equivariant. The shift equivariance properties would be lost with classical stridded convolutions \cite{cohen2016group}. This property is crucial for understanding the meaning of the elements in the symbolic representation. It ensures that identical symbolic elements always represent the same pattern (when there are no boundary effects), regardless of their position.

By changing the number of $B$ blocks in the encoder (and by the symmetry in the decoder), we can adjust the information learned in the representation, enabling us to adapt the representation to different frequency levels. The higher the number of blocks, the shorter the representation sequence and the more global the information captured for an element of the representation. Thus, we define the number of blocks $B$ as an hyperparameter of the architecture. 

\subsection{Training}
The unsupervised architecture is trained using the loss (eq. \ref{total_loss}) introduced by \cite{van2017neural}. To simplify the notations, we present the loss for an instance $\boldsymbol{x}$. The notations remain the same as those introduced above, except for the following. The operator $sg$ is the stop gradient operator whose derivative is zero during the backward computation time. The encoder is defined as $\phi_{\theta}$ and the decoder as $\psi_{\theta^{'}}$. In the Section \ref{Requirements}, $\phi_{\theta}$ characterizes the function that maps the time series to the symbolic representation. Here $\phi_{\theta}$ stands for the function that maps the time series to the embeddings (before the discretization phase). On the other hand, here, $\psi_{\theta^{'}}$ characterizes the function that maps the quantized vector representation to the reconstruction. The temporal vector sequence after quantization is denoted $\boldsymbol{E^{q}} \in \mathbb{R}^{T{'} \times Z}$. Thus, the total loss of the unsupervised model is:

\begin{align} 
\underset{{\theta,\theta',E}}{\text{arg~min}} ~  ||\boldsymbol{x}-\psi_{\theta^{'}}(\boldsymbol{E^{q}})||_{2}^{2} &+ ||sg[\phi_{\theta}(\boldsymbol{x})] - \boldsymbol{E^{q}}||_{2}^{2} \nonumber \\
&+ \beta ||\phi_{\theta}(\boldsymbol{x}) - sg[\boldsymbol{E^{q}}]||_{2}^{2}. 
\label{total_loss}
\end{align} 


The first term of the loss refers to the reconstruction ability of the model. In practice, we consider the mean square error pointwise between the real-time series and the decoder output. This part of the loss optimizes the decoder and the encoder. It's important to note that the \text{{arg~min}} operator in eq. \ref{vq_loss} is not differentiable at the VQ level. In \cite{van2017neural}, the authors suggest passing the gradients of the quantized vectors to their corresponding encoder vector outputs. This allows the reconstruction loss to optimize the encoder. The second term in the loss updates the used centroids $\boldsymbol{E^{q}}$ by moving them toward the embedding vectors $\phi_{\theta}(\boldsymbol{x})$ assigned to them. The centroids are initialized with a Gaussian distribution. This approach has proven to be stable in training and allows an independent choice of solver for the rest of the loss. In the model implementation, the Adam solver \cite{kingma2014adam} is used for the other terms of the loss. The last part of the loss is the commitment loss and ensures that the encoder outputs do not land too far from the used centroids. It guarantees better training stability. 

\section{Downstream task: classification over extracted representations}
\label{Classif_section}

Similar to dictionary methods for time series classification \cite{lin2003symbolic, schafer2012sfa}, the proposed representation learning process leads to an interpretable sequential symbolic representation of the time series. The series is represented by the sequence of the indices of the quantized vectors (the symbolic representation). We propose to evaluate the obtained symbolic representation with downstream classification tasks. When classifying symbolic representations, authors commonly extract histograms of symbolic subsequences and feed them into a K-NN, or linear classifier \cite{le2017time, le2019interpretable}. For example, in \cite{le2017time}, authors propose an efficient algorithm for a logistic regression classifier based on coordinate gradient descent to find the most relevant subsequences. In the following, we propose a linear classifier based on the presence or absence of discriminative subsequences inspired by existing methods. This classification method preserves the interpretability provided by the representation and ensures an in-situ, global interpretable classification decision.

\subsection{Classification using a unique symbolic representation}

In our classification process, we focus only on using subsequences of lengths one and two to characterize the symbolic representation. This approach effectively captures local dynamics while maintaining a low-dimensional feature space. Let $d$ be the cardinal of the set of possible symbolic subsequences of length one and two, and $\boldsymbol{r_i}$ be a symbolic representation (extracted using our auto-encoder) for individual i. We can extract for the representation $\boldsymbol{r_i}$, the vector $\boldsymbol{h_i}$ which indicates if a subsequence (of length one or two) is present in $\boldsymbol{r_i}$. Thus, $\boldsymbol{h_i}$ is a vector of size d composed of 0 and 1 elements ($\boldsymbol{h_i} \in \{0,1\}^{d}$). The features vectors $\{\boldsymbol{h_i}, i\in\{1\ldots N\} \}$ representing the training set can be used after that to solve the classification problem using logistic regression (eq. \ref{equation_classif}):
\begin{align}
    \underset{{w, b}}{\text{arg~min}} ~ &\frac{1-\rho}{2} \boldsymbol{w^{T}}\boldsymbol{w}+\rho\|\boldsymbol{w}\|_{1}+ \nonumber \\ 
    &\lambda \sum_{i=1}^{n} \log \left(\exp \left(-y_{i}\left(\boldsymbol{h_{i}}^{T} \boldsymbol{w}+b\right)\right)+1\right)
    \label{equation_classif}
\end{align}

 with $\lambda$ the regularization parameter, $\rho$ the trade-off between the $\ell_1$ penalty and the $\ell_2$ penalty and $b$ the bias. Sparsity in interpretable classification models is desirable to more easily understand the decision \cite{wu2018beyond, li2019exploiting}. In the case of multiclass classification, the usual One vs All method \cite{rifkin2004defense} is applied to provide better interpretability.
 
However, classification using a unique representation (e.g., a representation where the encoder/decoder is composed of B blocks) does not allow the use of features of different frequencies. Therefore, it is desirable to train several representations corresponding to different temporal dimension reductions (deeper or shallower architectures) in order to classify using different frequency features.

\subsection{Classification using multiple symbolic representations}
According to Subsection \ref{archi} the depth of the architecture is directly related to the number of blocks $B$ in the encoder (and by symmetry in the decoder). We train the proposed unsupervised model for $D$ different depths ($B \in \{1,2,...,D\}$). Thus, we get $D$ representations of different length, which capture different features $(\boldsymbol{h^{1}_i}, \boldsymbol{h^{2}_i}, ..., \boldsymbol{h^{D}_i})$. Then we apply the following steps:

\begin{enumerate}
    \item A penalized logistic regression (eq. \ref{equation_classif}) is performed separately on each extracted features vector $\boldsymbol{h_i}$.   
    \item For each $\boldsymbol{h_i}$, we recover the features whose logistic regression coefficients have non-zero values. Thus, we obtain an aggregated discriminative set of features for the D representations.  
    \item We perform a final penalized logistic regression on the obtained set.
\end{enumerate}
\label{process}

It is important to highlight two observations. Firstly, this classification method retains interpretability because each feature is interpretable by the construction of the representation, making it straightforward to interpret the coefficients of the logistic regression associated with a feature. Secondly, the number of representations we use corresponds to the number of times the initial time series can be halved while being greater than a threshold ($T'$ should be greater than this threshold). In practice, the number of representations we use for the UCR datasets rarely exceeds five. Keeping the number of representations small is essential for preserving interpretability. In particular, a large number of representations has a strong negative impact on interpretability \cite{DBLP:journals/cacm/Lipton18}.

\section{Experiments}
\label{sec:expe}

In our experiments, the unsupervised architectures are trained using only the training set. The symbolic test representations are constructed by passing the test data through the trained architecture. We also set the number of available centroids to 32 ($K=32$) and the dimension of the latent space is 64 ($Z=64$). For training the unsupervised models, we set $\beta$ to $0.25$ in the loss (eq. \ref{total_loss}) as realized in \cite{van2017neural}. 

\subsection{Quantitative experiments}
\label{quant}

The quantitative experiments are performed on 25 selected datasets from the UCR archive \cite{dau2019ucr}. These datasets meet specific criteria: variety of application types (sensor, motion, image, device), a minimum of 50 training and 50 test instances, and a maximum of seven classes. Table \ref{tab:results-interpretable} presents the results of our model compared to the interpretable in-situ methods SAX SEQL, SAX VSM, FS and LTS presented in Section \ref{RelatedContent}. We also compare our accuracy results to the well-known K-NN classifier coupled with Dynamical Time Warping (DTW) distance \cite{rakthanmanon2012searching}. This classifier is sometimes presented as interpretable, but it does not allow the extraction of localizable discriminative features. We use the accuracy metrics from \cite{le2019interpretable} for comparison. Referring to Table \ref{tab:results-interpretable}, our method gives better results than the other in-situ methods interpretable on average on these datasets. 

\begin{table}[!htb]
\setlength{\tabcolsep}{2pt}
\caption{Accuracy on 25 UCR datasets compare to in-situ interpretable methods. The best results are in bold and the second best results are underlined.}
\begin{tabular}{|l|c|c|c|c|c|c|}
\hline 
Datasets &  Ours &  \begin{tabular}[c]{@{}c@{}}SAX \\ SEQL\end{tabular} &  \begin{tabular}[c]{@{}c@{}}SAX \\ VSM\end{tabular} & FS &  \centering{LTS}  &  \begin{tabular}[c]{@{}c@{}}DTW \\ CV\end{tabular} \\
\hline 
Coffee                         &       0.964 &     \textbf{1.000} &   0.929 &  0.929 &             \textbf{1.000} &   \textbf{1.000} \\
Computers                      &       \textbf{0.728} &     \underline{0.676} &   0.620 &  0.500 &             0.584 &  0.620 \\
DistalPhalanxOAG               &       0.755 &     \underline{0.818} &   \textbf{0.842} &  0.655 &             0.779 &  0.626 \\
DistalPhalanxOC                &       \underline{0.732} &     0.718 &   0.728 &  \textbf{0.750} &             0.719 &  0.725 \\
DistalPhalanxTW                &       \underline{0.640} &     \textbf{0.748} &   0.604 &  0.626 &             0.626 &  0.633 \\
Earthquakes                    &       0.734 &     \textbf{0.789} &   \underline{0.748} &  0.705 &             0.741 &  0.727 \\
ECG5000                        &       \textbf{0.932} &     0.924 &   0.910 &  0.923 &             \textbf{0.932} &  0.925 \\
FordA                          &       \underline{0.883} &     0.851 &   0.827 &  0.787 &             \textbf{0.957} & 0.691 \\
GunPoint                       &       0.940 & \underline{0.987} & \underline{0.987} & 0.947 & \textbf{1.000} & 0.913 \\
Ham                            &       0.705 &     0.705 &   \textbf{0.810} &  0.648 &             0.667 &  0.600 \\
Herring                        &       \textbf{0.656} &     0.578 &   \underline{0.625} &  0.531 &             \underline{0.625} &  0.531 \\
ItalyPowerDemand               &       0.906 &     0.734 &   0.816 &  0.917 &             \textbf{0.970} &   \underline{0.955} \\
LargeKitchenApp         &       \underline{0.864} &     0.760 &   \textbf{0.877} &  0.560 &             0.701 &  0.795 \\
PhalangesOC                    &       \underline{0.748} &     0.717 &   0.710 &  0.744 &             \textbf{0.765} &   0.761 \\
ProximalPhalanxOC              &       0.818 &     0.818 &   \underline{0.828} &  0.804 &             \textbf{0.834} &  0.790 \\
ProximalPhalanxOAG             &       0.839 &     \underline{0.844} &   0.824 &  0.780 &             \textbf{0.849} &  0.785 \\
ProximalPhalanxTW              &       0.771 &     \textbf{0.792} &   0.610 &  0.702 &             \underline{0.776} &   0.756 \\
RefrigerationDevices           &       0.533 & \underline{0.541} & \textbf{0.653} & 0.333 & 0.515 & 0.440 \\
ScreenType                     &       \underline{0.499} &     0.461 &   \textbf{0.512} &  0.413 &             0.429 &   0.411 \\
ShapeletSim                    &       \underline{0.994}  & \underline{0.994} & 0.717 & \textbf{1.000} & 0.950 & 0.700 \\
SmallKitchenApp         &       \textbf{0.795} &     \underline{0.776} &   0.579 &  0.333 &             0.664 &  0.672 \\
Strawberry                     &       \textbf{0.962} &     0.954 &   \underline{0.957} &  0.903 &             0.911 &  0.946 \\
Wafer                          &       0.975 &     0.993 &   \textbf{0.999} &  \underline{0.997} &             0.996 &   0.995 \\
Wine                           &       \underline{0.759} &   0.556 &   \textbf{0.963} &  \underline{0.759} & 0.500 &  0.611 \\
Worms                          &       \textbf{0.714} &     0.536 &   0.558 &  \underline{0.649} &  0.610 &   0.532 \\
\hline 
\hline 
\textbf{Mean}     &       \textbf{0.793} & 0.770 & 0.769 & 0.715 & 0.764 & 0.725 \\
\hline 
\hline 
\end{tabular}
\label{tab:results-interpretable}
\end{table}

Our method results in a minimum 2.2 percentage points increase compared to other interpretable in-situ methods. Even if our method does not come first in all cases, the quantitative results are very promising for two main reasons. First, only a small subpart of all possible symbolic subsequences are used as our method aims to  enforce sparsity in order to favour interpretability. Secondly, our representations are learned without any supervision and thus are not guided by the underlined task.

In addition, we evaluate the accuracy of logistic regression on our symbolic neural representation against SVM on top of neural representation. This framework is often used to evaluate the best unsupervised neural representation when the neural representation is a simple vector \cite{DBLP:conf/aaai/YueWDYHTX22, franceschi2019unsupervised}. Previous research has found that the best-unsupervised methods with this framework are TS2Vec \cite{DBLP:conf/aaai/YueWDYHTX22} and T-Loss \cite{franceschi2019unsupervised}. We compare these two methods with our own on the 25 previous datasets and find that all three perform similarly in accuracy ($0.789 \pm 0.15$ for T-Loss and $0.807 \pm 0.15$ for TS2Vec compared to $0.793 \pm 0.13$ for our method). These results suggest that the constraints we impose on our architecture to satisfy the interpretability requirement do not significantly deteriorate the expressiveness of the unsupervised representation.

\subsection{Qualitative experiment}

Among the 25 previous UCR datasets, not all are suitable for interpretability because a limited number of features cannot discriminate between the classes. In most papers dealing with in-situ interpretability for time series, the commonly used datasets are GunPoint, ShapeletSim, or Coffee datasets \cite{le2017time, Senin2013}. For the qualitative analysis, we decide to focus on the GunPoint and ShapeletSim datasets \cite{dau2019ucr}.

\subsubsection{GunPoint dataset}

The GunPoint dataset consists of two actors performing a movement with their right hand, with two classes: Gun-Draw (class 0) and Point (class 1). 

\paragraph{Representations} 
The time series are z-normalized for each instance. We train five models with different architecture depths to capture different features in the time series. The architectures is only trained using the train time series, then a simple forward pass is used to get the representation for each time series in the test dataset. After training, we obtain five representations of different lengths. As shows in Table \ref{rep-info-table}, the representation generalizes easily to the test. 

\begin{table}[!htb]
\centering
\setlength{\tabcolsep}{2pt}
\caption{Information on the different representations learned for the \textit{GunPoint} dataset.}
\begin{tabular}{|c|c|c|c|c|}
\hline
\multicolumn{1}{|l|}{\textbf{Depth}} &
  \textbf{\begin{tabular}[c]{@{}c@{}}Temporal\\ downscaling\end{tabular}} &
  \textbf{\begin{tabular}[c]{@{}c@{}}Symbolic\\ representation\\ length\end{tabular}} &
  \textbf{\begin{tabular}[c]{@{}c@{}}Pointwise\\ train MAE\end{tabular}} &
  \textbf{\begin{tabular}[c]{@{}c@{}}Pointwise\\ test MAE\end{tabular}} \\ \hline
$B=1$ & $2^{1} = 2$  & 75 & 0.044 & 0.045 \\ \hline
$B=2$ & $2^{2} = 4$  & 38 & 0.038 & 0.045 \\ \hline
$B=3$ & $2^{3} = 8$  & 19 & 0.032 & 0.047 \\ \hline
$B=4$ & $2^{4} = 16$ & 10 & 0.027 & 0.085 \\ \hline
$B=5$ & $2^{5} = 32$ & 5  & 0.023 & 0.085 \\ \hline
\end{tabular}
\label{rep-info-table}
\end{table}

\paragraph{Classification}
For each trained representation $\boldsymbol{r_{B}}$ we construct the binary vector $\boldsymbol{h^{B}}$ ($B \in \{1,2,3,4,5\}$). We then fit a logistic regression for each representation separately. For each logistic regression (eq. \ref{equation_classif}), $\rho$ and $\lambda$ are found by cross-validation \cite{scikit-learn}. We encourage a strong penalty $\ell_1$ to set to zero the coefficients for non-discriminating features for each representation. Table \ref{TableReg} shows for each logistic regression the initial number of features and the number of features whose regression coefficient is different from zero. 

\begin{table}[!htb]
\centering
\setlength{\tabcolsep}{3pt}
\caption{Information on logistic regression for each representation (step 1 in process \ref{process}).}
\begin{tabular}{|c|c|c|}
\hline
\multicolumn{1}{|l|}{\textbf{Depth}} & \textbf{$\boldsymbol{h^{B}}$ size} & \textbf{\begin{tabular}[c]{@{}c@{}}Number of features actually \\ used to classify \end{tabular}} \\ \hline
$B=1$ & 115 & 23 \\ \hline
$B=2$ & 203 & 26 \\ \hline
$B=3$ & 293 & 24 \\ \hline
$B=4$ & 328 & 26 \\ \hline
$B=5$ & 187 & 11 \\ \hline
\end{tabular}
\label{TableReg}
\end{table}

Then, we fit the final logistic regression on each extracted feature (whose regression coefficients differ from 0). The final logistic regression performs on 110 features. After training, 89 coefficients are set to zero in this regression because of the penalty effect. Initially, the concatenation of feature vectors is a vector of size 1126, but only 21 features are ultimately used for the classification problem.  

The test accuracy is 0.94, and the train accuracy is 1. Now that the final logistic regression is fitted, we can look at the coefficients of this regression. Table \ref{classif-info-table} shows the most critical features (whose relative importance, their absolute value divided by the value of the others, is greater than 5 percent).

\begin{table}[!htb]
\centering
\setlength{\tabcolsep}{3pt}
\caption{Details of the most discriminative features in the final logistic regression (step 3 in process \ref{process}), intercept is 5.13.}
\begin{tabular}{|c|c|c|c|}
\hline
\multicolumn{1}{|l|}{\textbf{Depth}} &
  \textbf{\begin{tabular}[c]{@{}c@{}}Symbolic \\ subsequence\end{tabular}} &
  \textbf{\begin{tabular}[c]{@{}c@{}}Logistic regression \\ coefficients\end{tabular}} &
  \textbf{\begin{tabular}[c]{@{}c@{}}Relative\\ importance\end{tabular}} \\ \hline
$B=1$ & bb & - 0.92                      & 7.0 \% \\ \hline
$B=1$ & fc & - 0.93                      & 7.2 \% \\ \hline
$B=1$ & hk & - 1.20                      & 9.2 \% \\ \hline
$B=1$ & kg & - 0.98                      & 7.4 \% \\ \hline
$B=1$ & kh & - 0.76                      & 5.8 \% \\ \hline
$B=3$ & Fx & 1.23 & 9.4 \% \\ \hline
\end{tabular}
\label{classif-info-table}
\end{table}

Our representation is interpretable and allows us to decode symbolic subsequences, so we can use the extracted features and logistic regression coefficients to gain insight into the problem. We can interpret the classification decision at the global level as well as at the local level.

\begin{figure}[!htb]
    \centering
    \includegraphics[width=.90\linewidth]{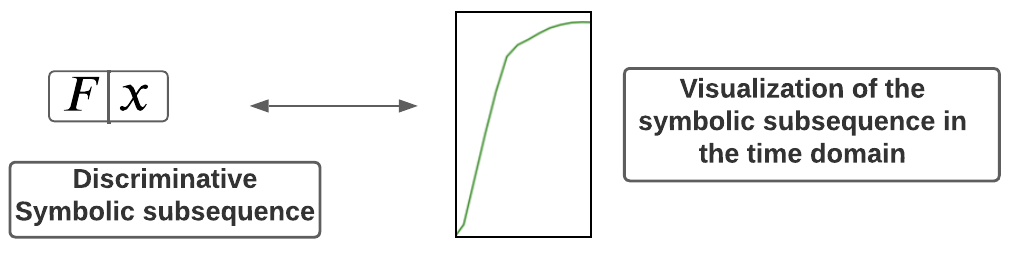}
    \caption{Global interpretability decision. Visualization of the most discriminative symbolic subsequence for class 1 (Point).}
    \label{fig:ShapeletVizu}
\end{figure}

\paragraph{Global interpretability} It consists in visualizing which feature at the model level allows the classification of the time series correctly. With our architecture, decoding the discriminative symbolic subsequences suffices to understand what the unsupervised model learned, and visualize the reconstructed subseries. Let us consider the symbolic sub-sequence \textit{'Fx'} (for depth $B=3$ in Table \ref{classif-info-table}). This sub-sequence is the most discriminative sub-sequence for class one. When we decode this subsequence in Figure \ref{fig:ShapeletVizu}, we obtain a subseries that characterizes the way the finger is raised. This subseries differs from the way the gun is raised. 

\begin{figure}[!htb]
    \centering
    \includegraphics[width=.99\linewidth]{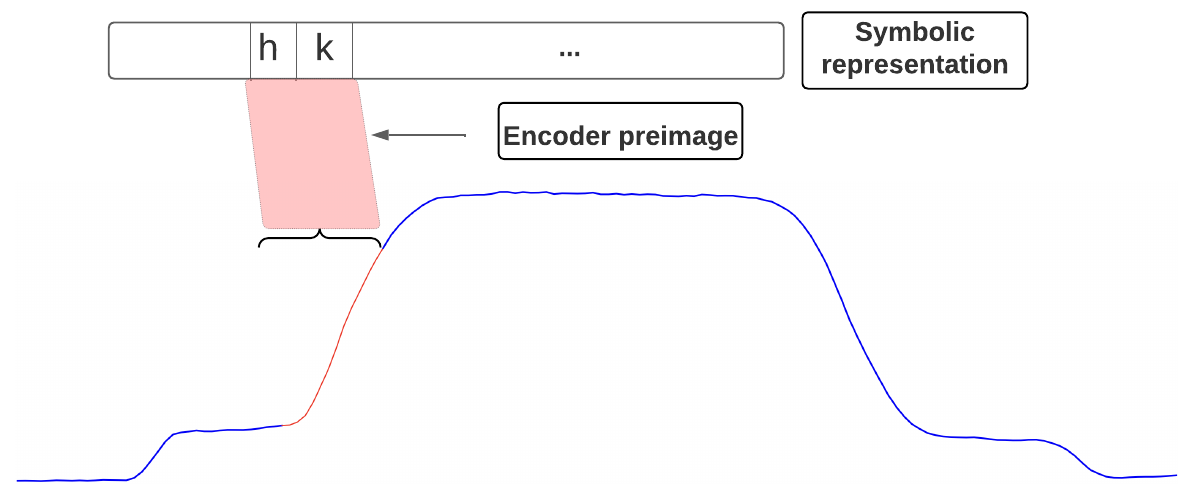}
    \caption{Local interpretability for a class 0 (Gun) instance.}
    \label{fig:localClass0}
\end{figure}

\paragraph{Local interpretability} It consists in visualizing for an instance the regions of the time series that make the decision. For example, take the symbolic subsequence \textit{'hk'} (for depth $B=1$ in Table \ref{classif-info-table}), which is the most discriminating subsequence for class 0. Figure \ref{fig:localClass0} presents an instance whose representation ($B=1$)  contains the subsequence \textit{'hk'}. Then, we highlight in red the pre-image of this subsequence using the receptive fields of the convolutions. 

\subsubsection{ShapeletSim dataset} The ShapeletSim dataset comprises two classes, class 0 is purely white noise, and class 1 includes a triangle shape at a random location. The classification problem is easily interpretable because only the presence or absence of a triangle characterizes the difference between the two classes.

\begin{figure}[!htb]
    \centering
    \includegraphics[width=0.90\linewidth]{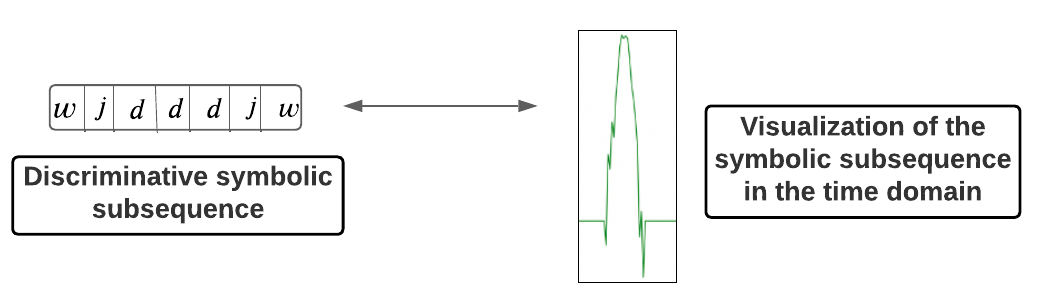}
    \caption{Global interpratibility decision. Visualization of the most discriminative symbolic subsequence for class 1 (includes triangle).}
    \label{fig:global}
\end{figure}

\begin{figure}[!htb]
    \centering
    \includegraphics[width=0.99\linewidth]{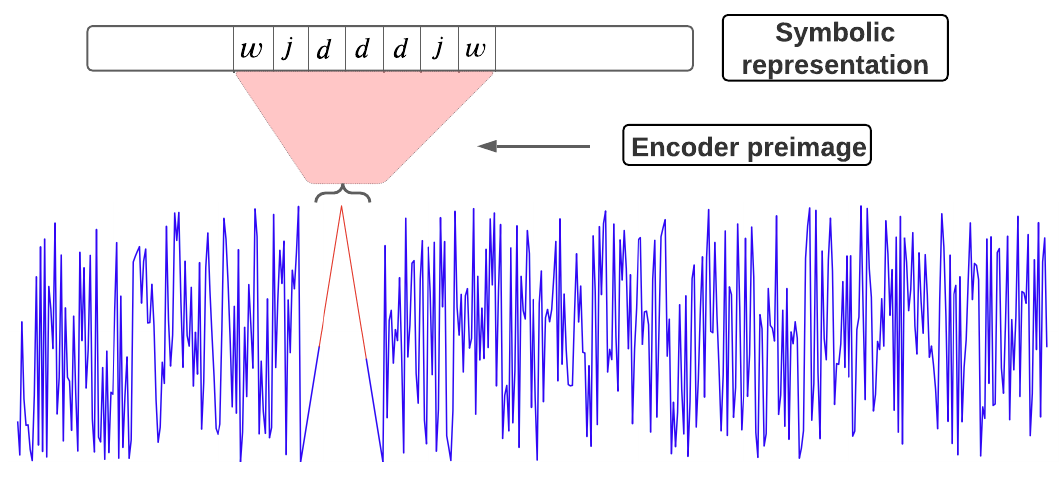}
    \caption{Local interpretability for a class 1 (includes triangle) instance.}
    \label{fig:local}
\end{figure}

As in the previous use case, we construct the different symbolic representations in an unsupervised manner and then extract the discriminative features. The highest coefficient in the final logistic regression is associated with the symbolic subsequence \textit{'wjdddjw'}. Figure \ref{fig:global} presents visualization for global interpretable classification decision. Using the decoder, we decoded the subsequence \textit{'wjdddjw'} and examined the resulting decoded shape. We observed that the decoded shape matches the vertex of the decoded triangle. It is worth noting that this demonstration of global interpretability does not require any specific instance for visualization.
On the other hand, Figure \ref{fig:local} shows a visualization of the local interpretability for the symbolic subsequence \textit{'wjdddjw'} for a given sample. We retrieve the triangular shape by computing the pre-image of the discriminating subsequence for this given sample. 

In the analysis of the ShapeletSim problem, we recognize its inherent simplicity, as indicated by the high test accuracy of 0.994 shown in Table \ref{tab:results-interpretable} when identifying the discriminant feature. However, it is an interesting dataset to visualize interpretability performance.
\section{Conclusion}
We first present essential requirements for building an interpretable neural representation for time series and then present an architecture that satisfies these requirements. The proposed unsupervised symbolic neural model fills a gap between symbolic and neural representations for time series. It has the advantage of allowing global interpretability of downstream classification tasks, while guaranteeing high expressiveness and good performance. The constructed representation has been evaluated both qualitatively and quantitatively on classification tasks. We show promising results for accuracy compared to both in-situ interpretable and neural methods. Additionally, the proposed interpretability provides an understanding of the model's classification decisions at both global and local levels for a broad range of time series. Much of our work has been devoted to the study of an unsupervised and interpretable neural architecture for time series. The choice of reconstruction loss in unsupervised architecture seems interesting for future work. Furthermore, coordinate descent, as described in \cite{ifrim2011bounded}, may improve the accuracy of these models by identifying longer symbolic subsequences that help distinguish between classes.

\section{Acknowledgment}

We would like to thank Tahar Nabil for the valuable discussions on this project.

\bibliographystyle{IEEEtran}
\bibliography{biblio}

\newpage

\appendices
\section{How to compute receptives fields regions}
\label{appendA}

With our model, it is very useful to be able to calculate the receptive fields and in particular the receptive field regions relative to an element (or a region) of the representation. To do this, we just need to adapt the following formulas computed in \cite{araujo2019computing} to our architecture: 

\begin{equation*}
v_{0}=v_{L} \prod_{i=1}^{L} s_{i}-\sum_{l=1}^{L}\left(1+p_{l}-k_{l}\right) \prod_{i=1}^{l-1} s_{i}
\end{equation*}
\begin{equation*}
u_{0}=u_{L} \prod_{i=1}^{L} s_{i}-\sum_{l=1}^{L} p_{l} \prod_{i=1}^{l-1} s_{i}
\end{equation*}

Where: 
\begin{itemize}
\item $v_0$ stands for the left-most coordinates of the receptive field in the intial time series
\item $u_0$ stands for the right-most coordinates of the receptive field in the intial time series
\item $v_L$ stands for the left-most coordinates in the representation
\item $u_L$ stands for the left-right coordinates in the representation
\item $k_l$ stands for the kernel size at depth l
\item $s_l$ stands for the stride at depth l
\item $p_l$ stands for the padding at depth l
\item L stands for the depth of the network
\end{itemize}

Thanks to these operations, we can highlight the receptive fields of the elements of the representation in Figure \ref{fig:local} and Figure \ref{fig:localClass0}.

\section{Impact on accuracy results of the number of available centroids (K)}
\label{appendB}

The quantitive experiments in Subsection \ref{quant} were performed for 32 centroids available during the vector quantization (K=32). It is interesting to see how the number of centroids affects the classification performance. Table \ref{tab:centroids} shows the accuracy results averaged over the 25 UCR datasets on which the experiments were conducted.

\begin{table}[h!]
\caption{Mean accuracy on the previous 25 UCR datasets for different k ($k \in \{8, 16, 32\}$). The best result are in bold and the second best result are underlined.}
\centering 
\begin{tabular}{|c||l|l|l|}
\hline
     & $k=8$  & $k=16$ & $k=32$                \\ \hline
\textbf{\begin{tabular}[c]{@{}c@{}}Mean \\ accuracy\end{tabular}} & \multicolumn{1}{c|}{0.753} & \multicolumn{1}{c|}{\underline{0.778}} & \multicolumn{1}{c|}{\textbf{0.793}} \\ \hline
\end{tabular}
\centering
\label{tab:centroids}
\end{table}

We observed that increasing the number of available centroids improves accuracy. However, it harms the centroids' expressiveness and, thus, the interpretability of the representation.

\section{Proof of the shift equivariance property}
\label{appendC}

In this appendix, we propose to describe how the adaptive polyphase downsampling/upsampling mechanism preserves the shift equivariance property despite downsampling (or upsampling). To do so, we pick up the proof given in \cite{DBLP:conf/acssc/ChamanD21}. We first present the proof of shift equivariance for the adaptive polyphase downsampling mechanism and then the adaptive polyphase upsampling case is straightforward.

\paragraph{Definition:} We define the shift equivariance property as follows. Let $S$ be an element of a sequence (a single element or a subsequence), and $G_{T}$ the group of discrete translations along the temporal axis. If we take $\tau$ to be any discrete translation in $G_{T}$ and $f$ to be a function equivariant by discrete translation for $G_{T}$, then there exists $\tau' \in G_{T'}$ such that: 
$f(\tau(S)) = \tau'(f(S))$.

\paragraph{Case 1: Adapative polyphase downsampling (APS-D) for downsampling by half}: 
To explain the APS-D operation, let us introduce $s(t)$ the convolutions output sequence of length $T_{0}$. We can then define the two sub-sequences $s_{0}$ and $s_{1}$ such that $s_{0}(t) = s(2t)$ and
$s_{1}(t) = s(2t+1)$. The APS-D operation $D^{A}_{2}$ consists in sub-sampling the sub-sequence $s_l$ such that: $$D^{A}_2(s) = s_l \quad \text{where} \quad l = \text{arg~max}{||s_j||_{1}} ~ (\text{for} ~ j \in \{0,1\}) $$

We introduce the following notations:
\begin{itemize}
    \item $\tau_{k}$ is a discrete translation by $k$
    \item $s_{APS} = D^{A}_2(s)$ is the non shift sequence output
    \item $s^{(k)}_{APS} = D^{A}_2(\tau_{k}(s))$ is the shifted sequence output
\end{itemize}

Then we have:
$$
s^{(k)}_{APS} = \begin{cases}
  \tau_{\frac{k}{2}}(s_{APS}), \text{when $k$ is even} \\
  \tau_{\frac{k+2i-1}{2}}(s_{APS}), \text{when $k$ is odd}
\end{cases}
$$

\paragraph{Case 2: Adapative polyphase usampling (APS-U) for upsampling by two:}
For the APS-U, the s signal is upsampled by two.  We insert 0 elements in each even or odd position according to the phase chosen in the block of the corresponding encoder. By construction, this operation is shift equivariant. For more details on this operation, see the original paper \cite{DBLP:conf/acssc/ChamanD21}.

\end{document}